%% file: main.tex
\documentclass{article}

\usepackage{deauthor}

\usepackage{latexsym}
\usepackage{graphicx}
\graphicspath{{./images/}}
\usepackage{booktabs} 
\usepackage{color}  
\usepackage{amsmath}  
\usepackage{subcaption}
\usepackage{caption}
\usepackage{tikz}
\usepackage{colortbl} 
\usepackage{framed}
\usepackage{multirow}
\usepackage{multicol}
\usepackage{hyperref}
\usepackage{url}
\usepackage{balance}
\usepackage{verbatim}
\usepackage{cancel}
\usepackage{xspace} 
\usepackage[ruled,vlined]{algorithm2e}
\usepackage{bbold} 
\usepackage{balance}
\usepackage{stmaryrd}
\usepackage{enumitem}
\usepackage{array} 
\usepackage{bold-extra} 

\newcommand{\utterance}[1]{\textit{#1}}
\newcommand{\phrase}[1]{\textit{``#1''}}

\newenvironment{Snugshade}[1][236,236,236]{
    \setlength{\itemsep}{0pt}
     \setlength{\parsep}{0pt}
     \setlength{\topsep}{0pt}
     \setlength{\partopsep}{0pt}
     \setlength{\leftmargin}{1.5em}
     \setlength{\labelwidth}{0em}
     \setlength{\labelsep}{0em} 
    \setlength{\parskip}{0pt}
    \definecolor{shadecolor}{RGB}{#1}%
    \begin{snugshade}
}{%
    \end{snugshade}%
}

\newcommand{\method}{\textsc{Quasar}\xspace}

\newcommand{\convinse}{\textsc{Convinse}\xspace}
\newcommand{\explaignn}{\textsc{Explaignn}\xspace}
\newcommand{\clocq}{\textsc{Clocq}\xspace}
\newcommand{\unikqa}{\textsc{UniK-Qa}\xspace}
\newcommand{\gptthree}{\textsc{Gpt-3}\xspace}
\newcommand{\gptfour}{\textsc{Gpt-4}\xspace}
\newcommand{\llama}{\textsc{Llama3}\xspace}
\newcommand{\spaghetti}{\textsc{Spaghetti}\xspace}

\newcommand{\compmix}{\textsc{CompMix}\xspace}
\newcommand{\timequestions}{\textsc{TimeQuestions}\xspace}
\newcommand{\crag}{\textsc{Crag}\xspace}

\newcommand{\squishlist}{
    \begin{list}{$\bullet$}{
        \setlength{\itemsep}{0pt}
	\setlength{\parsep}{3pt}
	\setlength{\topsep}{3pt}
	\setlength{\partopsep}{0pt}
	\setlength{\leftmargin}{1.5em}
	\setlength{\labelwidth}{1em}
	\setlength{\labelsep}{0.5em}
    }
}

\newcommand{\squishend}{
    \end{list}
}

\newcommand{\myparagraph}[1]{\vspace*{0.2cm}\noindent \textbf{#1}.}
\newcommand{\myparagraphnospace}[1]{\noindent \textbf{#1}.}

\newcommand{\GW}[1]{\emph{{\color{blue} GW:#1}}}

\begin{document}

\title{RAG-based Question Answering \\ over Heterogeneous Data and Text}

\author{
Philipp Christmann,
Gerhard Weikum\\\\
Max Planck Institute for Informatics\\
Saarland Informatics Campus, Germany\\
\texttt{\{pchristm, weikum\}@mpi-inf.mpg.de}}

\maketitle

\input{sections/00-abstract}
\input{sections/01-intro}

\input{sections/02-background}

\input{sections/03-method}
\input{sections/04-experiments}
\input{sections/05-discussion}

\input{sections/10-bib}

\end{document}

%% file: sections/00-abstract.tex
\section*{Abstract}
This article presents the \method system for question answering over unstructured text, structured tables, and knowledge graphs, with unified treatment of all sources.
The system adopts a RAG-based architecture, with a pipeline of evidence retrieval followed by answer generation, with the latter powered by a 
moderate-sized
language model.
Additionally and uniquely, \method
has components for question understanding, to derive crisper input for evidence retrieval, and for re-ranking and filtering the retrieved evidence before feeding the most informative pieces into the answer generation.
Experiments with three different benchmarks demonstrate the high answering quality of our approach, being on par with or better than large GPT models, while keeping the computational cost and energy consumption orders of magnitude lower.

%% file: sections/01-intro.tex
\label{sec:intro}
\section{Introduction}


\noindent\textbf{Motivation and Problem.} The task of question answering, QA for short, arises in many flavors: factual vs. opinions, simple lookups vs. multi-hop inference, single answer vs. list of entities, 
direct answers vs. long-form, one-shot questions vs. conversations, and other varieties 
(see, e.g., surveys \cite{RogersGA:CS2023,RoyAnand:MC2021}).
The state-of-the-art for this entire spectrum has been greatly advanced in the past decade. Most notably, incorporating deep learning into retriever-reader architectures (e.g., \cite{DBLP:conf/acl/ChenFWB17,DBLP:conf/eacl/IzacardG21,DBLP:conf/emnlp/KarpukhinOMLWEC20}) has boosted answering quality, and most recently, large language models (LLM) \cite{Minaee-LLM-survey,Zhao-LLMsurvey} have pushed the envelope even further (e.g., \cite{DBLP:journals/arXiv/abs-2305-06984}).

Today’s LLMs alone are capable of accurately answering many \textit{factoid} questions, simply from their pre-trained parametric memory which latently encodes huge text corpora and other online contents.
However, this critically depends on the frequency of evidence in the underlying contents and the complexity of the information need. 
For example, 
asking for the {\em MVP of the 2024 NBA season} would easily return the correct answer Nikola Jokic, 
but asking for the {\em highest-scoring German NBA player} or the {\em MVP of the 2024 German basketball league} pose a big challenge.
The reason is that LLMs alone do not easily recall information about not so popular or even long-tail entities \cite{Kandpal:ICML2023,Sun:NAACL2024},
and that they are mainly geared for direct look-ups as opposed to connecting multiple pieces of evidence \cite{Mavi:FnT2024,Zhang:NAACL2024}.

Supervised fine-tuning, often with human instructions, and the recent paradigm of retrieval-augmented generation 
\cite{DBLP:journals/arXiv/abs-2312-10997,Guu-REALM:ICML2020,DBLP:conf/nips/LewisPPPKGKLYR020,Zhao:arxiv2024}
known as RAG, address these bottlenecks. In addition to cleverly crafted prompts and few-shot examples, the LLM is provided with the top-ranked results of an explicit retrieval step, like web search or knowledge graph (KG) lookups. The former is often necessary for freshness of answers, and the latter may help with long-tail entities and also mitigate the notorious risk of hallucinations. Still, this generation’s RAG architectures are limited in how broad and how deep they tap into external sources. Popular AI assistants like Gemini or ChatGPT seem to primarily retrieve from the text of web pages (incl. Wikipedia articles), and academic research has additionally pursued knowledge augmentation by enhancing prompts with facts from large KGs (e.g., Wikidata).

An additional content modality that is still underexplored are {\em online tables}: a wide range of tabular data including HTML tables in web pages, spreadsheets and statistics, all the way to CSV and JSON datasets that are abundant on the Internet. There is prior work on joint support for text and KGs and for text and tables, but very little on all of these together -- some notable exceptions being \cite{Christmann-CONVINSE:SIGIR2022,Christmann-Explaignn:SIGIR2023,Oguz-UniK-QA:NAACL2022,Zhang-Spaghetti:ACL2024}.


\noindent\textbf{Examples.} All three heterogeneous types of sources are crucial not only for answering different questions from different kinds of evidence, but also for combining multiple pieces of evidence of different modalities to infer correct and complete answers.
To 
illustrate
the need for tapping all sources, consider the following questions:

\vspace*{0.2cm}
\begin{Snugshade}
$Q1$: \utterance{Which Chinese basketballers have played in the NBA?}\\
 \indent $Q2$: \utterance{Who was the first Chinese NBA player?}\\
  \indent $Q3$: \utterance{Which Chinese NBA player has the most matches?}
\end{Snugshade}
\vspace*{0.2cm}

Q1 can be cast into querying a KG, but the list there is not necessarily complete and up-to-date, so additional evidence from text or tables would be desired. 
Q2 needs information about who played in which seasons, found only in web pages or sports-statistics tables. 
Finally, Q3 may be lucky in finding prominent textual evidence (e.g., in biographies, Wikipedia etc.), but this often faces divergent statements, and resolving contradictions needs to dive into more evidence. Besides, when textual evidence is rare and hard to find or not trustworthy enough, then information from multiple tables and text snippets may have to be aggregated (e.g., totals of per-season counts).
Some of this may perhaps become feasible for an industrial LLM’s RAG capabilities in the near future, but there are always harder scenarios by moving from Chinese NBA players deeper into the long tail, such as asking for {\em Lithuanian players in the German handball league}.

\vspace*{0.2cm}
\noindent\textbf{Approach and Contribution.} This paper presents a simple but powerful and versatile RAG system with unified access to text, KG and tables. We call our method {\em \method} 
(for Question Answering over Heterogeneous Sources with Augmented Retrieval).
Its architecture is relatively straightforward: all heterogeneous content is verbalized and indexed for retrieval; a retriever finds top-ranked results for the given question (from different source types), and these are fed into the LLM for answer generation. This is the unsurprising bird-eye’s view. Specific details that are key factors for the strong performance of \method are: 

\squishlist
\item[i)] automatically casting user questions into a structured representation of the information need, which is then used to guide 
\item[ii)] judicious ranking of search results, with multiple rounds of re-ranking and pruning, followed by
\item[iii)]	extracting faithful answers from an LLM in RAG mode, with answers grounded in tangible evidence.
\squishend


\vspace{0.2cm}
\noindent The paper presents experiments with three different benchmarks, covering various flavors of questions.
We focus on one-shot questions; conversational QA is out of scope here, but \method itself is well applicable to this case, too.
Our experiments demonstrate that our methods are competitive, on par with big GPT models and often better,
while being several orders of magnitude lower in computational and energy cost.
The experimental findings also highlight that question understanding, with structured representation of user intents, and iterative re-ranking of evidence are crucial for good performance.

Overall, our contribution lies in proposing a unified system architecture for RAG-based question answering over a suite of different data sources, with strong points regarding both effectiveness (i.e., answer quality)
and efficiency (i.e., computational cost).


%% file: sections/02-background.tex
\label{sec:background}
\section{Related Work}


The RAG paradigm came up as a principled way of enhancing LLM factuality incl. provenance and mitigating the risk of hallucination \cite{Guu-REALM:ICML2020, DBLP:conf/nips/LewisPPPKGKLYR020}.
It is highly related to the earlier
retriever-reader architectures for QA \cite{DBLP:conf/acl/ChenFWB17,DBLP:conf/emnlp/KarpukhinOMLWEC20}, especially when the reader uses the fusion-in-decoder method \cite{DBLP:conf/eacl/IzacardG21,Oguz-UniK-QA:NAACL2022}.
Since its invention, RAG methodology has been greatly advanced, introducing a wide suite of extensions, such as batched inputs, interleaving retrieval and generation steps, and more (see the recent surveys \cite{DBLP:journals/arXiv/abs-2312-10997,Zhao:arxiv2024}).

On question answering (QA), there is a vast amount of literature including a wealth of differently flavored benchmarks (see, e.g., \cite{RogersGA:CS2023}).
The case of interest here is QA over heterogeneous sources, tapping into both unstructured content and structured data. 
A variety of works has pursued this theme by combining knowledge graphs with text sources, using graph-based methods, neural learning and  language models (e.g., \cite{Pramanik-Uniqorn:JWS2024,Sun-PullNet:EMNLP2019,Yasunaga:NAACL2021}).
%

Most relevant for this article is the research on jointly leveraging all different sources: text, KGs, and tables (incl. CSV and JSON files). This includes  
the \unikqa system~\cite{Oguz-UniK-QA:NAACL2022},
the \spaghetti/SUQL project \cite{Liu-SUQL:NAACL2024,Zhang-Spaghetti:ACL2024},
the \textsc{Matter} method~\cite{Lee-MATTER:ACL2024},
the STaRK benchmarking~\cite{Wu-STARK:arxiv2024},
and our own prior work
 \cite{Christmann-CONVINSE:SIGIR2022,Christmann-Explaignn:SIGIR2023}
 (without claiming exhaustiveness).
 Out of these, we include \unikqa, \spaghetti and our own systems \convinse and \explaignn as baselines in the experimental evaluation.
Their architectures are similar to ours, but \unikqa and \spaghetti do not have our distinctive elements of
question understanding and iterative re-ranking (originally introduced in \explaignn~\cite{Christmann-Explaignn:SIGIR2023}).


%% file: sections/03-method.tex
\label{sec:method}
\section{Methodology}

\begin{figure}[tb]
  \includegraphics[width=\textwidth]{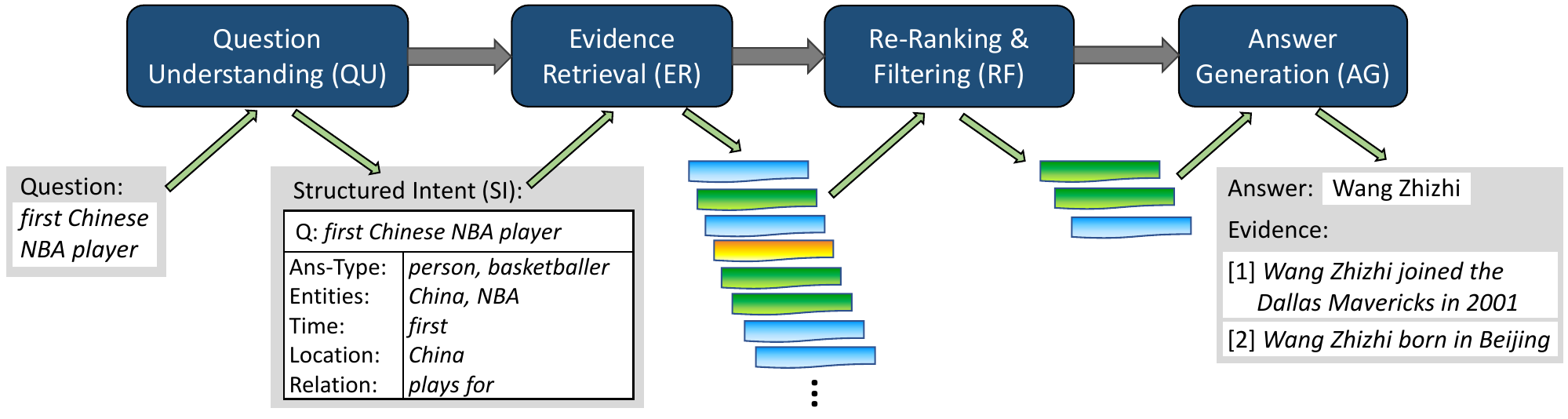}
  \caption{Overview of the \method system.}
  \label{fig:compass-overview}
\end{figure}

The \method system is a pipeline of four major stages, as illustrated in Figure \ref{fig:compass-overview}.
First, the input question is analyzed and decomposed, in order to compute a {\em structured intent (SI)} representation that will pass on to the subsequent steps, along with the original question. Second, the SI is utilized to retrieve pieces of evidence from different sources: text, KG and tables. 
Third, this pool of potentially useful evidence is filtered down, with iterative re-ranking, to arrive at a tractably small set of most promising evidence.
The final stage generates the answer from this evidence,
passing back the answer as well as evidence snippets for user-comprehensible explanation.

The second and fourth stage, Evidence Retrieval (ER) and Answer Generation (AG), are fairly standard. Such a two-phase architecture was called a retriever-reader architecture~\cite{Zhu-ODQA-survey:arxiv2021}. With a modern LLM replacing the earlier kinds of neural readers, this is the core of every RAG system~\cite{DBLP:journals/arXiv/abs-2312-10997}.

Stages 1 and 3 are unique elements of our architecture, judiciously introduced to improve both effectiveness (i.e., answer quality) and efficiency (i.e., computational cost).
Question Understanding (QU) provides the ER component with crisper and semantically refined input, and
the Re-Ranking \& Filtering (RF) stage is beneficial for
distilling the best evidence from the large pool of retrieved pieces.
The following subsections elaborate on the four stages of the pipeline, emphasizing the \method-specific steps QU and RF.

\subsection{Question Understanding (QU)}

To prepare the retrieval from different kinds of sources, including a KG, ad-hoc tables and text documents, it is useful to analyze and decompose the user question.
In this work, we aim to cast a question into a 
{\em structured intent (SI)} representation: essentially
a frame with faceted cues as slots, or equivalently, a concise set of key-value pairs. 
Figure \ref{fig:compass-overview} gives an idealized example for the question about the first Chinese NBA player. The facets or keys of potential interest here 
are:
\squishlist
\item {\em Ans-Type:} the expected answer type (or types when considering
different levels of semantic refinement), 
\item {\em Entities:} the salient entities in the question, and 
\item {\em Relation:} phrases that indicate which relation (between Q and A entities) the user is interested in. 
\squishend
\noindent In addition, as questions can have temporal or spatial aspects, the SI also foresees slots for:
\squishlist
\item {\em Time:} cues about answer-relevant time points or spans, including relative cues (e.g., ``before Covid'') and ordinal cues (e.g., ``first''), and
\item {\em Location:} cues about answer-relevant geo-locations.
\squishend

\vspace{0.2cm}
\noindent The ideal SI for example question Q2 would look like:

\begin{Snugshade}
{\em Ans-Type:} person, basketballer; {\em Entities:} China, NBA; {\em Time:} first;
{\em Location:} China; {\em Relation:} plays for.
\end{Snugshade}

Note that the values for these slots can be crisp like entity names or dates, but they can also take the form of surface phrases. The SI purpose and value lie in the decomposition. In practice, many questions would only lead to a subset of faceted cues, leaving some slots empty. For the example in Figure \ref{fig:compass-overview}, an alternative SI could simply consist of

\begin{Snugshade}
{\em Ans-Type:} person; {\em Entities:} China, NBA; {\em Time:} first.
\end{Snugshade}

\noindent Even this simplified SI can be highly beneficial in guiding the subsequent evidence retrieval.

To generate the SI from a user question, we employ a (small-scale) LM, specifically BART~\cite{DBLP:conf/acl/LewisLGGMLSZ20}, a Transformer-based auto-encoder with 140M parameters.\footnote{\url{https://huggingface.co/facebook/bart-base}}
BART is pre-trained for language representation; its power for our purpose comes from fine-tuning.
To this end, we generate (question, SI) pairs by using an instruction-trained LLM like GPT-4, with few-shot in-context learning (following our earlier work~\cite{Jia-FAITH:WWW2024}). 
Note that this is a one-time action; at inference-time we only use much smaller LMs.
The generated silver-standard pairs are then used to fine-tune BART.
In the experiments in this article, we leverage pre-existing collections of silver pairs, based on the training data of the CompMix benchmark~\cite{Christmann-CompMix:WWW2024}, 
comprising $3{,}400$ such pairs.

Although this paper focuses on single-shot questions, the \method architecture is also geared for conversational QA. In that setting, the SI can play an even bigger role, as (follow-up) questions are often formulated in a rather sloppy manner -- all but self-contained. For example, a conversation could start with a clear question {\em When did Wang Zhizhi join the NBA?}, followed a few dialog steps later, by a user utterance like {\em Which teams did he play for?} or simply {\em Which teams?}.
In such an informal conversation, the system needs to {\em contextualize} each user utterance based on the preceding turns in the dialog (e.g., inferring the relevant entities Wang Zhizhi and NBA from the conversational history).
For details on conversational QA, based on our architecture, see our earlier works~\cite{Christmann-CONVINSE:SIGIR2022,Christmann-Explaignn:SIGIR2023}.

\subsection{Evidence Retrieval (ER)}

The ER stage taps into a knowledge graph, a corpus of text documents, and a collection of web tables.
Specifically, for the experiments, we use the Wikidata KG,
all English Wikipedia articles, and all tables that are embedded in Wikipedia pages (incl. infoboxes, which can be seen as a special case of tables). 

\vspace{0.2cm}
\noindent{\bf Retrieval from KG:}
To retrieve evidence from the KG, we utilize our earlier work
\clocq~\cite{Christmann-CLOCQ:WSDM2022}, which provides entity disambiguations and a relevant KG-subgraph for a given query.
Unlike most other works on QA-over-KG, \clocq fetches all KG-facts that are relevant for a given entity in a single step.
For example, when querying for
NBA players, it can traverse the KG neighborhood and pick up top teams, also considering so-called qualifier nodes in Wikidata which are often used for temporal scopes. 
As the disambiguation of entity names onto the KG can be tricky and noisy (e.g., China could be mapped to Chinese sports teams in all kinds of sports), \clocq considers several possible disambiguations~\cite{Christmann-CLOCQ:WSDM2022} (typically in the order of $10$ result entities).
The queries for \clocq are 
constructed by concatenating all slots of the question's SI.
For the example query about the first Chinese NBA player,
good result entities would be Dallas Mavericks, lists about NBA seasons, MVP awards etc., and their associated facts. These provide cues, but are likely insufficient to answer the question.

\vspace{0.2cm}
\noindent{\bf Retrieval from Text and Tables:}
The disambiguated entities returned by \clocq form anchors for tapping into text and tables.
\method first identifies 
relevant text documents and tables that refer to the anchor entities. With focus on Wikipedia, these are simply the articles for the respective entities. 
\method then constructs a keyword query that concatenates all available fields of the SI.
The query is evaluated against a linearized and verbalized representation (see below) of all sentences and all table rows in the selected documents.
This returns a set of sentences and 
and individual table rows, ranked by BM25 scores.


\vspace{0.2cm}
\noindent{\bf Evidence Verbalization:}
All results from the different data sources are uniformly treated by {\em linearizing} and {\em verbalizing} them
into token sequences. For KG results, the entity-centric triple sets are linearized via breadth-first traversal of the mini-graph starting from the entity node.
For tables, results are individual rows, which are contextualized by including labels from column headers and from the DOM-tree path of the article where the table comes from. For example, a table row about Wang Zhizhi playing for Dallas (Mavericks) in the 2000-2001 season, would be expressed as:

\vspace{0.05cm}
\hspace*{0.5cm} Wang Zhizhi / NBA Career / Season: 2000-2001, Team: Dallas, Games Played: 5 \dots
\vspace{0.05cm}

\noindent Finally, results from the text corpus are already in the form of token sequences, but we can additionally prefix these with the DOM-tree labels.
We can think of this entire pool of evidence as 
an on-the-fly corpus of potentially relevant pseudo-sentences, forming the input of the subsequent RF stage.

\vspace{0.2cm}
\noindent {\bf Result Ranking:}
Overall, the ER stage compiles a substantial set of evidence, possibly many thousands of entities, text snippets and table rows. Therefore, we practically restrict the pool to a subset of high-scoring pieces, like the top-$1000$.
For scoring, a simple BM25 model (a classical IR method) is applied. 
By default, we treat all evidence pieces uniformly with global scoring, no matter whether they come from KG, text or tables. 


\subsection{Re-Ranking and Filtering (RF)}

With a pool of top-$1000$ evidence pieces, we could invoke an LLM for answer generation. However, that would face a large fraction of noise (i.e., misleading evidence) and incur high costs of computation and energy consumption. 

For both of these reasons, we have devised light-weight techniques for iteratively reducing the top-$1000$ pieces to a small subset, say top-$30$ or top-$10$, that can be fed into an LLM at much lower cost (as LLM computations and pricing are at least linear in the number of input tokens). The difficulty is, of course, to do this without losing good evidence and reducing answer presence. Our techniques for this task are based on graph neural networks (GNNs)~\cite{Wu:IEEE2021} or cross-encoders (CEs)~\cite{Dejean:arxiv2024,Lin:MC2021}.

\myparagraph{GNN-based RF}
Given a large pool of evidence pieces from all sources, a bipartite graph is constructed:
\squishlist
\item {\em nodes} being evidence pieces or entities that occur in these pieces, and
\item {\em edges} connecting an evidence piece and an entity if the entity occurs in the evidence.
\squishend

The task for the GNN is to jointly score the evidence and the entity nodes in a multi-task learning setup. The latter are the {\em answer candidates}, and the evidence should give {\em faithful explanation} for an answer.
We build on our earlier work on explainable QA~\cite{Christmann-Explaignn:SIGIR2023}.

The node encodings are initialized with cross-encoder embeddings (see below) 
for node contents and the SI of the question. The inference iteratively adjusts the encodings based on message passing from neighboring nodes.
The GNN is trained via weak supervision from question-answer pairs:
evidence nodes are labeled as relevant if they are connected to
a gold answer.
More technical details are given in~\cite{Christmann-Explaignn:SIGIR2023}.

\method invokes the GNN in multiple rounds, iteratively reducing top-$k$ to top-$k^*$ nodes with $k^* \ll k$. In practice, we would typically consider two rounds: re-ranking top-$1000$ and pruning to top-100, and then reducing to top-30 or top-10, which are passed to the answer generation stage.
Note that this keeps the GNN at a tightly controlled size, so that its computational costs at inference-time are much smaller than those of an LLM.

\myparagraph{CE-based RF}
An alternative to the GNN inference is to employ a cross-encoder for scoring and re-ranking the evidence pieces.
These are transformers (typically with a small LM like BERT) that are fine-tuned for scoring the relatedness between a query and a document~\cite{Nogueira:arxiv2019}. In our case, the comparison is between the question SI and the evidence piece. In our experiments, we make use of two different cross-encoders, 
both trained on the MS-MARCO benchmark for passage retrieval~\cite{Bajaj:arxiv2018}, 
and fine-tuned on the respective benchmark (leveraging the same weak supervision data as for the GNNs),
the difference being in model size.\footnote{\url{https://huggingface.co/cross-encoder/ms-marco-MiniLM-L-4-v2} and\\ \url{https://huggingface.co/cross-encoder/ms-marco-MiniLM-L-6-v2}}
%
%
We use the smaller model to reduce top-$1000$ to top-100, and the larger model to go further down from top-100 to top-30.


\subsection{Answer Generation (AG)}

The last stage follows mainstream practice to invoke an LLM in a retrieval-augmented manner.
We call a `small-scale` LLM, specifically a fine-tuned LlaMA-3.1 model (8B-Instruct)\footnote{\url{https://huggingface.co/meta-llama/Llama-3.1-8B-Instruct}}, with a prompt \footnote{The specific prompt is \phrase{SI: \textless\texttt{concatenated SI}\textgreater \hspace{0.1cm} Evidence: \textless\texttt{evidence pieces}\textgreater}.}
consisting of:

\squishlist
\item the concatenated SI of the original question, and
\item the top-30 (or other top-$k^*$ with small $k^*$) evidence pieces.
\squishend

By the previous down-filtering of the original pool of evidence pieces, this last step has affordable cost in terms of computation time and energy consumption.

\vspace{0.2cm}
\noindent{\bf Fine-Tuning the LLM:}
We considered adding an instruction to the prompting, such as {\em ``answer this question solely based on the provided evidence snippets''}.
However, this turned out to be ineffective.
The reason why the model works well without such instructions is our task-specific fine-tuning.
We perform this by running the training data of benchmarks through the \method pipeline,
and training the AG stage with the top-30 evidence pieces as input.
Thus, the fine-tuning makes the model learn the role of evidence for RAG-based QA.

\vspace{0.2cm}
\noindent{\bf Explanations:}
The top-30 evidence pieces can be used to provide users with explanation of answers.
Optionally, these could be reduced further for comprehensibility.
Alternatively, we can fine-tune the LLM to provide both answers and concise explanations.
Since we can infer which evidences in the input mention the annotated ground-truth answers,
our method could be fine-tuned to provide such \textit{answering evidences} as well (cf. \cite{Gao-citations:emnlp2023}).


%% file: sections/04-experiments.tex
\label{sec:exp}
\section{Experiments}

\label{setup}
\subsection{Experimental setup}

\myparagraphnospace{Benchmarks} We run experiments on three benchmarks with different characteristics of questions.

\squishlist
    \item \textbf{\compmix}.
    \compmix~\cite{Christmann-CompMix:WWW2024} is a benchmark which was specifically designed for evaluating QA systems operating over heterogeneous sources. The dataset has $9{,}410$ questions, out of which $2{,}764$ are used for testing.
    Answers are crisp entity names, dates, or other literals.
    
    \item \textbf{\crag}.
    We further evaluate on a subset of the \crag~\cite{Yang-CRAG} dataset, which was recently released as a testbed for RAG-based QA systems.
    We utilize the same pipeline and sources as outlined in Section~\ref{sec:method}, without using the web snippets or APIs provided with \crag. This way we focus on entity-centric questions that do not require access to live web data (e.g., news feeds), and disregard cases where the results would be up-to-date quantities.
    This restricts the test data to $436$ entity-centric questions, still enough for a proof of concept.
    
    \item \textbf{\timequestions}.
    To showcase the generalizability of our pipeline, we conduct experiments on~\timequestions~\cite{Jia-TimeQuestions},
    a benchmark for temporal QA. The dataset requires temporal understanding and reasoning, which are well-known limitations of
    LLMs~\cite{Dhingra-time-aware-LLM:TACL2022}. \timequestions has 16{,}181 questions (3{,}237 for testing).
\squishend

Typical examples for the questions in these three benchmarks are:

\begin{Snugshade}
\compmix: \utterance{Which player won the most number of Man-of-the-Match titles in the FIFA world cup of 2006?}\\
 \indent \crag: \utterance{What was the worldwide box office sales for little hercules?}\\ 
  \indent \timequestions: \utterance{Which club did Cristiano Ronaldo play for before joining Real Madrid?}
\end{Snugshade}

\myparagraph{Baselines} As competitors or reference points to \method, we study the performance of the following methods:

\squishlist
    \item \textbf{Generative LLMs}.
    We compare \method against out-of-the-box LLMs: \textbf{\gptthree} (\texttt{text-davinci-003}), \textbf{\gptfour} (\texttt{gpt-4}) 
    and \textbf{\llama} (\texttt{meta-llama/Llama-3.1-8B-Instruct}).
    The same prompt is used for all LLMs, consistent with previous work~\cite{Christmann-CompMix:WWW2024, Zhang-Spaghetti:ACL2024}:
    \phrase{Please answer the following question by providing the crisp answer entity, date, year, or numeric number. Q: \textless\texttt{question}\textgreater}.

    \item \textbf{Heterogeneous QA methods}.
    \convinse~\cite{Christmann-CONVINSE:SIGIR2022}, \unikqa~\cite{Oguz-UniK-QA:NAACL2022}, \explaignn~\cite{Christmann-Explaignn:SIGIR2023}
    are QA methods designed to integrate heterogeneous sources: text, tables and KG. All of these  integrate the exact same sources as \method.

    \item \textbf{\textsc{State-of-the-art}}.
    For \compmix and \timequestions, we also compare against state-of-the-art methods from the literature: \spaghetti~\cite{Zhang-Spaghetti:ACL2024} and \textsc{Un-Faith}~\cite{Jia-FAITH:WWW2024}, which are among the best performing systems.
    
Results are taken from the literature whenever applicable.
On \crag, we use the models trained on \compmix for \method and heterogeneous QA baselines.
\squishend

\myparagraph{Metrics}
We measure \textit{precision at 1} (\textbf{P@1}) as our main metric~\cite{RoyAnand:MC2021} on all benchmarks.
On \crag, we manually annotate answer correctness, as the ground-truth answer formats vary (e.g., entity name variants, lists, sentences).

We also compute the number of neural parameters aggregated over all sub-modules (\textbf{\#Parameters}).
Parameter counts for GPT-models are taken from~\cite{Minaee-LLM-survey}
(\gptfour might have less active parameters during inference).

For further analysis we measure \textit{answer presence} (\textbf{AP@k}),
i.e. whether the answer is present in the top-$k$ ranked evidence pieces,
and \textit{mean reciprocal rank} within the top-$k$ evidences (\textbf{MRR@k}).

\myparagraph{Configuration}
Our implementation uses the \texttt{Llama3.1-8B-Instruct} model for the AG stage.
For the QU, ER and RF stages
we adopt code from the \explaignn project.\footnote{\url{https://explaignn.mpi-inf.mpg.de}}
For the ER stage, we use \clocq, setting its specific parameters to $k=10$ and $p=1{,}000$.

As default, we use the GNN technique for the RF stage.
For efficiency, we use light-weight models for initializing
the GNN encoders -- the same models used for the CE-based RF.\footnote{\url{https://huggingface.co/cross-encoder/ms-marco-MiniLM-L-4-v2} and\\\url{https://huggingface.co/cross-encoder/ms-marco-MiniLM-L-6-v2}}
The GNNs are trained for $5$ epochs with an epoch-wise evaluation strategy,
i.e. we choose the model with the best performance on the respective dev set.
We train the GNNs on graphs with a maximum of $100$ evidence and $400$ entity nodes (as scored by BM25).
During inference, the first GNN is applied on graphs with $1{,}000$ evidence and $4{,}000$ entity nodes, shrinking the pool of evidence pieces to the top-$100$.
The second GNN then runs on graphs with $100$ evidence and $400$ entity nodes.
The factor of 4 entities per evidence (on average) holds sufficient for the observed data,
and enables batched inference.
Other parameters are kept as is.

The AG model, based on \texttt{Llama3.1-8B-Instruct}, is 
fine-tuned
for $2$ epochs with a warm-up ratio of $0.01$ and a batch size of $8$, again with an epoch-wise evaluation strategy.
Other parameters are set to the default Hugging Face
training parameters.\footnote{\url{https://huggingface.co/docs/transformers/v4.46.2/en/main_classes/trainer\#transformers.TrainingArguments}}

\subsection{Main results}
\myparagraphnospace{\method is competitive on all benchmarks}
Main results of our experiments are shown in Table~\ref{tab:main-res}.
First of all, we note that \method achieves competitive performance across all three benchmarks.

On \compmix, baselines for heterogeneous QA and \llama perform similarly,
whereas GPT-based LLMs can answer more than $50$\% of the questions correctly.
\method exhibits substantially higher performance, on par with
the state-of-the-art method \textsc{Spaghetti}~\cite{Zhang-Spaghetti:ACL2024}
(which is based on \gptfour).

On the \crag dataset, P@1 drops for all methods except for \gptfour. 
The benchmark includes realistic questions,
which can be ambiguous/confusing (\phrase{who was the director for the report?}),
on ``exotic'' entities with answers in social media (\phrase{how many members does the teknoist have?}),
or require up-to-date information (\phrase{when did chris brown release a song or album the last time?}),
and other cases that are challenging for all methods.

Finally, \method establishes new state-of-the-art performance on the \timequestions benchmark.
Interestingly, all of the tested LLMs show greatly reduced performance on this benchmark,
which inherently requires temporal understanding and reasoning
-- a known weakness of stand-alone LLMs.


\begin{table} [h]
    \centering
    \newcolumntype{G}{>{\columncolor [gray] {0.90}}c}
    \begin{tabular}{l G G G c}
        \toprule
            \textbf{Method $\downarrow$ / Benchmark $\rightarrow$} & \textbf{\compmix}  & \textbf{\crag} & \textbf{\timequestions} & \textbf{\#Parameters} \\ 
        \midrule
            \textbf{\gptthree} 
            & $0.502$ &   $-$ & $0.224$ & $175{,}000$ M  \\ 

            \textbf{\gptfour}
            & $0.528$ &   $\mathbf{0.633}$ & $0.306$ & $1{,}760{,}000$ M  \\ 

            \textbf{\llama~\cite{Touvron-LLaMA}} (8B-Instruct)
            & $0.431$ &   $0.385$ & $0.178$ & $8{,}030$ M  \\ 

        \midrule
            \textbf{\convinse~\cite{Christmann-CONVINSE:SIGIR2022}}
            & $0.407$  &   $0.298$ & $0.423$  & $362$ M \\ 

            \textbf{\unikqa~\cite{Oguz-UniK-QA:NAACL2022}}
            & $0.440$ &   $0.280$ & $0.424$ & $223$ M  \\ 
    
            \textbf{\explaignn~\cite{Christmann-Explaignn:SIGIR2023}} 
            & $0.442$ &   $0.303$ & $0.525$  & $328$ M \\ 

        \midrule 
            \textbf{\textsc{State-of-the-art}}
            & $\mathbf{0.565}$ 
            & $-$
            & $0.571$  & $-$ \\

            & (\textsc{Spaghetti}~\cite{Zhang-Spaghetti:ACL2024})
            & 
            & (\textsc{Un-Faith}~\cite{Jia-FAITH:WWW2024}) &  \\

        \midrule
            \textbf{\method (ours)}
            & ${0.564}$ &   $0.362$ & $\mathbf{0.754}$  & $8{,}218$ M \\ 
        \bottomrule
    \end{tabular} 
    \vspace*{-0.2cm}
    \caption{End-to-end P@1 of \method and baselines on three benchmarks. Results for \gptthree and \gptfour are taken from the literature~\cite{Christmann-CompMix:WWW2024, Jia-FAITH:WWW2024}. \gptthree is not accessible anymore, hence no results on \crag.
    }
    \label{tab:main-res}
\end{table}

\myparagraph{Integration of heterogeneous sources is vital}
\method integrates evidence from text, KG and tables into a unified framework.
We aim to better understand how this affects the answering performance of the method.
Table~\ref{tab:sources} shows end-to-end answering performance of \method
with different combinations of the input sources.
The results clearly indicate that all types of sources contribute, with option Text+KG+Tables performing best,
with a large margin over tapping only single source types.


\begin{table} [t] 
    \centering
    \newcolumntype{G}{>{\columncolor [gray] {0.90}}c}
    \newcolumntype{H}{>{\setbox0=\hbox\bgroup}c<{\egroup}@{}}
    	\begin{tabular}{l G G G H H H c c c} 
        \toprule
            \textbf{Benchmark $\rightarrow$}
                & \multicolumn{3}{G}{\textbf{\compmix}} 
                & \multicolumn{3}{H}{\textbf{\crag}}
                & \multicolumn{3}{c}{\textbf{\timequestions}} \\ 
        \midrule
            \textbf{Input sources $\downarrow$ / Metric $\rightarrow$}
                & \textbf{P@1} & \textbf{AP@100}  & \textbf{AP@30}
                & \textbf{P@1} & \textbf{AP@100}  & \textbf{AP@30}
                & \textbf{P@1} & \textbf{AP@100}  & \textbf{AP@30} \\
            \midrule
                \textbf{Text}           &  $0.455$  &  $0.563$  &  $0.531$ &  $?$  &  $?$  &  $?$ &  $0.539$  &  $0.515$  &  $0.487$   \\
                \textbf{KG}             &  $0.481$  &  $0.677$  &  $0.637$ &  $?$  &  $?$  &  $?$ &  $0.724$  &  $0.701$  &  $0.674$   \\
                \textbf{Tables}         &  $0.432$  &  $0.501$  &  $0.482$ &  $?$  &  $?$  &  $?$ &  $0.536$  &  $0.347$  &  $0.328$   \\
            \midrule
                \textbf{Text+KG}        &  $0.537$  &  $0.749$  &  $0.706$ &  $?$  &  $?$  &  $?$ &  $0.745$  &  $\mathbf{0.776}$  &  $0.748$   \\
                \textbf{Text+Tables}    &  $0.503$  &  $0.632$  &  $0.594$ &  $?$  &  $?$  &  $?$ &  $0.567$  &  $0.578$  &  $0.549$   \\
                \textbf{KG+Tables}      &  $0.524$  &  $0.728$  &  $0.692$ &  $?$  &  $?$  &  $?$ &  $ 0.743$  &  $0.731$  &  $0.703$   \\
            \midrule
                \textbf{Text+KG+Tables}    &  $\mathbf{0.564}$  &  $\mathbf{0.759}$  &  $\mathbf{0.724}$ &  $?$ &  $?$  &  $?$ & $\mathbf{0.754}$    & $\mathbf{0.776}$  &  $\mathbf{0.749}$   \\
            \bottomrule
    \end{tabular}
    \vspace*{-0.2cm}
    \caption{Answer presence and answering precision of \method with different combinations of input sources (on the respective test sets).}
    \label{tab:sources}
\end{table}

\subsection{Analysis}


\myparagraph{Unified retrieval enhances performance}
In the RF stage, we re-rank and filter evidence from different source types,
and feed the unified top-\textit{k}* into the AG stage.
We conduct a comparison in which we consider
the top-$10$ evidence pieces from each source type individually. This gives equal influence to KG, text and tables, whereas our default is based on global ranking.
Table~\ref{tab:unified-retrieval} shows the results for this analysis, showing our default choice performs better.
The reason is that different questions require different amounts of evidence from each of the source types.

\begin{table} [t] 
    \centering
    \newcolumntype{G}{>{\columncolor [gray] {0.90}}c}
    \newcolumntype{H}{>{\setbox0=\hbox\bgroup}c<{\egroup}@{}}
    	\begin{tabular}{l G G H H} 
        \toprule
            \textbf{Input evidences $\downarrow$ / Metric $\rightarrow$} & \textbf{P@1} & \textbf{AP@30}  & \textbf{\crag} & \textbf{\timequestions} \\ 
            \midrule
                \textbf{Top-30 Text+KG+Tables (ours)}             &  $\mathbf{0.574}$  &  $\mathbf{0.710}$  &  $-$  &  $-$   \\
                \textbf{Top-10 Text + Top-10 KG + Top-10 Tables}        &  $0.560$    &  $0.709$  &  $-$  &  $-$   \\
            \bottomrule
    \end{tabular}
    \vspace*{-0.2cm}
    \caption{Answer presence and precision
    of \method for different choices of top-30 
    (on \compmix dev set).}
    \label{tab:unified-retrieval}
\end{table}

\begin{figure}[t]
    \centering
    \includegraphics[width=0.8\textwidth]{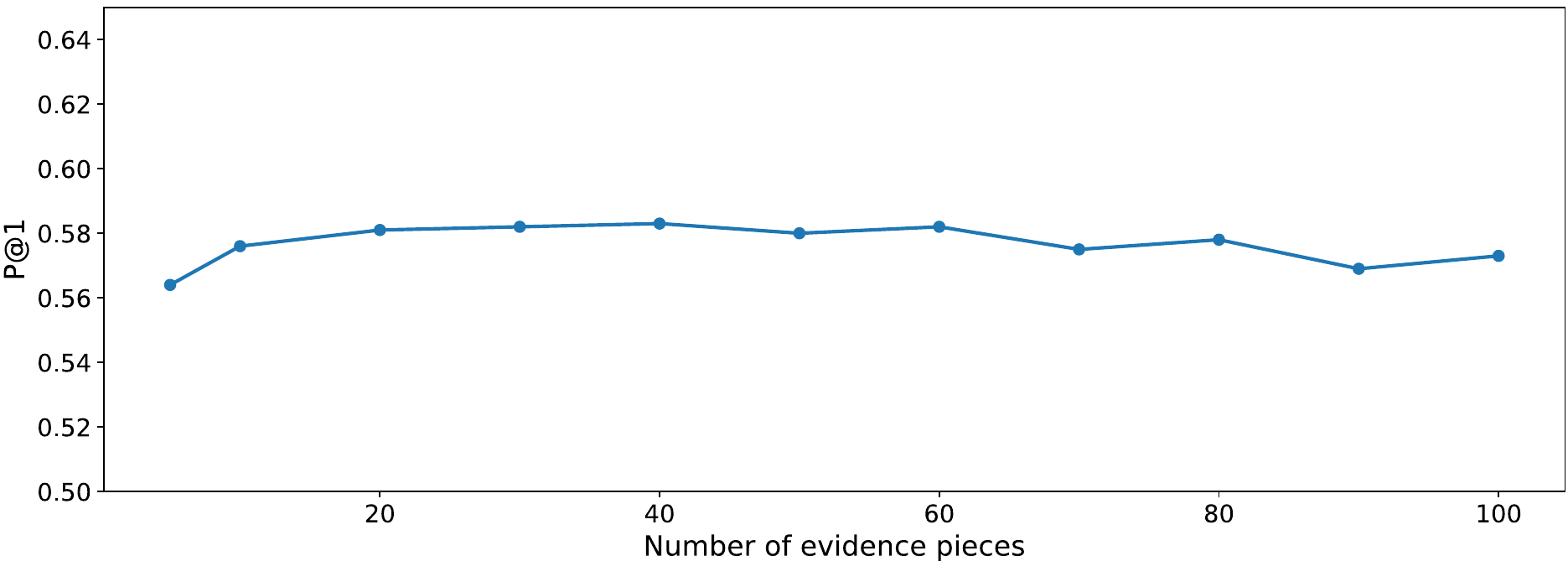}
    \vspace*{-0.2cm}
    \caption{Performance of \method on the \compmix dev set with different numbers of evidence.}
    \label{fig:res-num-evidences}
    \vspace*{-0.2cm}
\end{figure}

\myparagraph{\method works well with small amounts of evidence}
We investigate the 
influence of 
the number of evidence pieces
fed into the AG stage, varying it from $5$ to $100$.
Results are shown in Figure~\ref{fig:res-num-evidences}.
As the curve shows, there is a sharp increase in precision as we add evidence up to 30 or 40 pieces, which is around our default of top-30. This indicates that a certain amount of evidence is needed, to overcome the inherent noise and arrive at sufficient answer presence. 
As we increase the amount of evidence further, we observe a saturation effect, and eventually a degration of performance. Too much evidence not only has diminishing returns, but can actually be confusing for the AG stage. This reconfirms our heuristic choice of top-30: enough for good answering while keeping computational costs reasonably low.


\myparagraph{Ablation study on re-ranking} For more insight on the possible configurations of the RF stage, we conducted an ablation study with different options, including solely relying on the initial BM25 scoring without explicit re-ranking. The results are shown in Table \ref{tab:ablation2}. We observe that the iterative reduction in two steps is slightly better than the single-step variants (going down from top-1000 to top-30 in one RF step). Between the two options of using a GNN or a CE, the differences are negligible. A notable effect is that our RF techniques retain the answer presence at a very high level, only a bit lower than for the initial top-1000. 
The last two rows of Table \ref{tab:ablation2} demonstrate that RF is crucial: without explicit re-ranking, the technique of just picking smaller top-$k$ from the original BM25 model leads to substantial degradation in both answer presence and precision.

\begin{table} [t] \small
    \centering
    \newcolumntype{G}{>{\columncolor [gray] {0.90}}c}
    \newcolumntype{H}{>{\setbox0=\hbox\bgroup}c<{\egroup}@{}}
    	\begin{tabular}{l G H G G G} 
        \toprule
            & \multicolumn{5}{G}{\textbf{\compmix} (dev set)} \\
        \midrule
            \textbf{RF Method $\downarrow$ / Metric $\rightarrow$} & \textbf{P@1} & \textbf{AP@1000} & \textbf{AP@100}  & \textbf{AP@30} & \textbf{MRR@100} \\ 
        \midrule
            \textbf{GNN: 1000 $\rightarrow$ 100 $\rightarrow$ 30}             &  $\mathbf{0.574}$  & $0.760$ &  $0.738$  &  $0.710$  &  $\mathbf{0.572}$  \\
            \textbf{CE: 1000 $\rightarrow$ 100 $\rightarrow$ 30}          &  $0.573$  & $0.760$  &   $\mathbf{0.740}$ &  $\mathbf{0.721}$   &  $0.553$   \\
        \midrule
            \textbf{GNN: 1000 $\rightarrow$ 30} &  $0.567$  & $?$  &   $n/a$ &  $0.710$   &  $0.567$   \\
         \textbf{CE: 1000 $\rightarrow$ 30} &  $0.570$  & $0.760$  &   $n/a$ &  $0.715$   &  $0.558$   \\
            \midrule
           \textbf{BM25: 100 (w/o GNN or CE)} &  $0.490$ & $0.760$  &  $0.652$  &  $n/a$  &  $0.259$ \\       
            \textbf{BM25: 30 (w/o GNN or CE)} &  $0.468$  & $0.760$  &  $n/a$ &  $0.534$   &  $0.259$   \\
            \bottomrule
    \end{tabular}
    \vspace*{-0.2cm}
    \caption{Ablation study for different RF strategies of \method on the \compmix dev set. The answer presence in the RF input with top-$1000$ evidence pieces is $0.760$.}
    \label{tab:ablation2}
\end{table}



\myparagraph{Quality of SI}
To assess the quality and robustness of the Structured Intents, we
inspected a sample of questions and their SIs.
Table~\ref{tab:question-SI-examples} gives three anecdotic examples.
We show SIs generated by \method, which makes use of the pre-existing collection from the \compmix benchmark for training.
This training data was obtained via different heuristics, 
which can be a limiting factor when user intents become more complex.

Therefore, we also looked at SIs derived via in-context learning (ICL) using \gptfour with $5$ handcrafted examples.
As shown in our earlier work on temporal QA~\cite{Jia-FAITH:WWW2024},
such data can be used for training smaller models (e.g., BART),
which can greatly boost the completeness and overall quality of the generated SIs.

From the sampled set, we observed that the ICL-based SIs are more
complete with all slots filled, whereas the BART-based SIs focused more
on the main slots Answer-type, Entities and Relation.
However, both approaches achieve very high quality in filling the slots,
capturing the user's information need very well.

Interestingly, when questions get complicated, with nested phrases, 
the ICL-based variant succeeds in decomposing the questions, based on only $5$ ICL examples.
For example, for the question {\em ``which German state had the most Corona-related death cases in the first year after the outbreak?''}
the Time slot becomes {\em ``first year after Corona outbreak''},
which can be resolved to identify the temporal scope.
In general, we believe that such question decomposition, beyond simple temporal constraints,
would be an interesting theme for future work.

\begin{table} [t] 
    \centering
    \small
    \newcolumntype{G}{>{\columncolor [gray] {0.90}}c}
    \newcolumntype{H}{>{\setbox0=\hbox\bgroup}c<{\egroup}@{}}
\resizebox*{\textwidth}{!}{
        \begin{tabular}{p{6cm}|p{6cm}|p{6cm}} 
        \toprule
            \textbf{Question} & \textbf{Current SI by \method} & \textbf{SI via ICL} \\
        \midrule
        \textit{what was disneys first color movie?}
            & Ans-Type: \textit{animated feature film} & Ans-Type: \textit{film, animated film} \\
            & Entities: \textit{disneys} & Entities: \textit{Disney} \\
            & Relation: \textit{was first color movie} & Relation: \textit{first color movie} \\
            &  & Time: \textit{first} \\
        \midrule
        \textit{at the oscars, who won best actor in 2018?}
            & Ans-Type: \textit{human} & Ans-Type: \textit{person, actor} \\
            & Entities: \textit{at the oscars} & Entities: \textit{Oscars, 2018} \\
            & Relation: \textit{who won best actor in 2018} & Relation: \textit{won best actor} \\
            &  & Time: \textit{2018} \\
        \midrule
        \textit{which German state had the most Corona-} & Ans-Type: \textit{state} & Ans-Type: \textit{location, state} \\
        \textit{related death cases in the first year after} & Entities: \textit{Germany, Corona} & Entities: \textit{Germany, Corona-related deaths} \\
        \textit{the outbreak?} & Relation: \textit{which state had the most related} & Relation: \textit{highest count of death cases} \\
        & \textit{death cases in the first year after the out-}  & Location: \textit{Germany} \\
            & \textit{break}  & Time: \textit{first year after Corona outbreak} \\
        \bottomrule
    \end{tabular}
}
    \vspace*{-0.2cm}
    \caption{Examples for pairs of question and generated SI.}
    \label{tab:question-SI-examples}
\end{table}

\myparagraph{Refraining from answering}
We can train our model to refrain from answering in scenarios
where the provided evidence does not contain an answer to the question.
Specifically, during training, when the answer is not present in the evidence,
we change the target answer to {\em unknown}. This variant is referred to as \method {\em (faithful)}.

We measure the ratio of questions for which {\em unknown} is provided as answer,
and the P@1 restricted to questions that are answered.
The accuracy of refraining from answering is measured as well,
based on whether the answer is present in the evidence or not.
We conduct this experiment on \compmix and \timequestions,
for which we can compute answer presence exactly.
We also compute results for \llama, which is already instructed 
with the option to answer ``don't know''.
Table~\ref{tab:refrain-from-answer} shows the results.
For \compmix, we observe that \method has high accuracy on refraining when appropriate,
whereas \llama tends to be overconfident with a very small rate of {\em unknowns}, leading to incorrect answers.

\begin{table} [t] \small
    \centering
    \newcolumntype{G}{>{\columncolor [gray] {0.90}}c}
    \newcolumntype{H}{>{\setbox0=\hbox\bgroup}c<{\egroup}@{}}
    	\begin{tabular}{l G G G G c c c c} 
        \toprule
            & \multicolumn{4}{G}{\textbf{\compmix}} & \multicolumn{4}{c}{\textbf{\timequestions}} \\
            \midrule
            \textbf{Metric $\rightarrow$} & \textbf{P@1}  & \textbf{P@1} & \textbf{Refrain} & \textbf{Refrain} & \textbf{P@1}  & \textbf{P@1} & \textbf{Refrain} & \textbf{Refrain} \\ 
            \textbf{Method $\downarrow$} &                 & \textbf{(answered)} & \textbf{rate} & \textbf{accuracy} &                 & \textbf{(answered)} & \textbf{rate} & \textbf{accuracy} \\ 
            \midrule
                \textbf{\llama}             &  $0.431$  &  $0.471$  &  $0.089$  &  $n/a$  &  $0.177$  &  $0.276$  &  $0.392$  &  $n/a$  \\
                \textbf{\method (faithful)}           &  $0.497$  &  $0.713$  &  $0.303$  &  $0.838$ &  $0.597$  &  $0.804$  &  $0.257$  &  $0.864$  \\
            \bottomrule
    \end{tabular}
    \vspace*{-0.2cm}
    \caption{
        Performance of \method with option to refrain from answering (``don't know'').
    }
    \label{tab:refrain-from-answer}
\end{table}

%% file: sections/05-discussion.tex
\label{sec:disc}
\section{Insights, Limitations, and Challenges}


\noindent{\bf Benchmark Performance.} Our method, RAG-based \method with an 8B LLaMA model, outperforms much larger LLMs like \gptfour on two of the three benchmarks, with a very large margin for temporal questions. Obviously, pre-trained LLMs have only limited sense of properly positioning ``remembered’’ facts on the timeline even with training data that exceeds ours by several orders of magnitude. This confirms our intuition that LLMs alone are not good at ``recalling’’  higher-arity relations that require combining distant pieces of evidence. This is a sweet spot for RAG. Only for 
the \crag benchmark, \method is substantially inferior to a full-blown LLM. This is likely due to the nature of the questions: not necessarily the complexity of the information needs, but the need for more web sources (beyond what our experiments tap into).

\vspace{0.2cm}
\noindent{\bf Cost/Performance Ratio.} The most important take-away from our experiments is that \method achieves its competitive performance at a much lower cost than the full LLMs. Assuming that the consumed GFlops are proportional to the number of model parameters, \method achieves a cost reduction by a factor of 200x for \gptthree and 2000x for \gptfour. This does not only mean less computation, but also a massively lower electricity bill and climate impact.  

\vspace{0.2cm}
\noindent{\bf Role of Question Understanding.} We did not systematically investigate the influence of the Structured Intent in the \method pipeline. However, the comparison to the big GPT models reflects the role of the SI, as we prompt the GPT models in their natural mode with the original questions. The linearized sequence of available SI slots does not always have major advantages, but there are enough cases where specific facets provide crucial cues. This holds especially for the Entities slot, as this drives the gathering of evidence in the ER stage (cf. \cite{Christmann-CONVINSE:SIGIR2022}, and for the Time slot, as these cues are often decisive for temporal questions (cf. \cite{Jia-FAITH:WWW2024}).

\vspace{0.2cm}
\noindent{\bf Role of Re-Ranking.} As our ablation studies show, merely using top-$k$ evidence from an initial BM25-style ranking does not provide good performance. Also, there seems to be sweet spot in the choice of $k$: we need enough evidence for connecting the dots if the question requires multiple pieces of information, or for corroborating candidates if the question finds many useful but noisy pieces. In the experiments, $k=30$ turns out to be good choice; much lower $k$ results in insufficient evidence, and much larger $k$ leads to saturation and ultimately degrading performance. Our argument for iteratively shrinking the candidate set in multiple rounds of re-ranking is substantiated in our experiments, but the gain of doing this, compared to GNN- or CE-based re-ranking from 1000 to 30, is not big. More research is called to better understand the role of ranking in RAG. 

\vspace{0.2cm}
\noindent{\bf Limitations of Evidence Retrieval.}
For ER, we adopted more or less standard techniques. The results showed very good answer presence, in the order of 75\% in the top-100 or even top-30. An important case where this is insufficient are questions that require aggregating information over a large number of evidence pieces. An example is asking for the life-time total of 3-point scores of the basketball player Dirk Nowitzki.
This requires collecting a set of per-season tables with NBA player statistics, but also other web sources with numbers for his career before he joined the NBA (including his youth teams).
Of course, there are sometimes shortcuts like a Wikipedia article or biography mentioning the total number, but this cannot be universally assumed. The bottom line is that ER should be reconsidered as well, striving to improve the recall dimension.

\vspace{0.2cm}
\noindent{\bf Limitations of Answer Generation.}
For AG, we simply rely on a LLM,
using it as an extractor (``reader'') from the given evidence. Despite the wide belief that LLMs can perform deep
reasoning over many pieces of evidence, our experience is that the extraction works only well – robustly and faithfully – for relatively simple questions with a few multi-hop joins or simple aggregation over a few pieces. However, complicated questions such as asking for the top-100 NBA players with the largest number of life-time 3-point scores (again including their pre-NBA careers) are currently out of scope and will likely remain so for quite some time. This offers many opportunities for pushing the envelope further.

\vspace{0.2cm}
\noindent{\bf Trust in Data Sources.}
In our experiments, we considered all heterogeneous sources as trustworthy and unbiased. With focus on Wikidata and Wikipedia, this assumption has been well justified. In the wild, however, input data for RAG-based systems likely exhibit a wide spectrum of quality issues, in terms of stale information, biased positions, or simply false statements. Identifying trustworthy and up-to-date evidence and dealing with conflicting data, has been explored in other contexts (e.g., for KG curation \cite{Dong-Trust:PVLDB2015}), but remains a major challenge for RAG-based QA.

\vspace{0.2cm}
\noindent{\bf Open Challenges and Future Work.} The best-performing methods in our experiment, mostly \method, reach P@1 values of 56\% for \compmix and 75\% for \timequestions. 
For the latter, the answer presence in the top-100 is only slightly higher; so the AG stage hardly misses anything.
However, for \compmix, the answer presence is 75\% -- much higher than what our system can actually answer. Obviously, closing this gap is a major direction to pursue, with focus on the RF and AG stages. However, missing one fourth of the answers completely in the top-100 pool, is a big problem as well. This requires improving recall at the ER stage, possibly with better guidance by the QU, which in turn needs more sources beyond the scope of our experiments (currently limited to Wikidata and Wikipedia). 

In general, we need to think beyond this kind of ``benchmark mindset’’. Even if we reached 80\% or 90\% precision and recall, we would still have a substantial fraction of questions that are answered incorrectly
or not at all. 
The remaining errors may not be a problem for chatbots, but they would be a showstopper for the deployment of mission-critical applications in business or science. We believe that this big gap is a shortcoming of {\em all methods}, not an issue that comes from the data alone. For trivia-style QA, as looked at in this paper, a smart human in ``open book’’ mode and no time limitation should be able to properly answer practically all questions, just by reading pieces of web contents and putting things together. Neither LLMs nor state-of-the-art RAG are the final solution; substantial research and creative ideas are needed to further advance QA.

%% file: sections/10-bib.tex

\newcommand{\bibauthors}[1]{{#1}}
\newcommand{\bibtitle}[1]{\emph{#1}}
\newcommand{\bibconf}[1]{{#1}}

%% file: main.bbl
\vspace*{1cm}
\begin{thebibliography}{10}

\bibitem{Bajaj:arxiv2018}
\bibauthors{Payal Bajaj, Daniel Campos, Nick Craswell, Li Deng, Jianfeng Gao, Xiaodong Liu, Rangan Majumder, Andrew McNamara, Bhaskar Mitra, Tri Nguyen, Mir Rosenberg, Xia Song, Alina Stoica, Saurabh Tiwary, Tong Wang.}
\bibtitle{MS MARCO: A Human Generated MAchine Reading COmprehension Dataset.}
In \bibconf{arXiv 2018}.

\bibitem{DBLP:conf/acl/ChenFWB17}
\bibauthors{Danqi Chen, Adam Fisch, Jason Weston and Antoine Bordes.}
\bibtitle{Reading Wikipedia to Answer Open-Domain Questions.}
In \bibconf{ACL 2017}.

\bibitem{Christmann-CONVINSE:SIGIR2022}
\bibauthors{Philipp Christmann, Rishiraj Saha Roy, Gerhard Weikum.}
\bibtitle{Conversational Question Answering on Heterogeneous Sources.}
In \bibconf{SIGIR 2022}.

\bibitem{Christmann-CLOCQ:WSDM2022}
\bibauthors{Philipp Christmann, Rishiraj Saha Roy, Gerhard Weikum.}
\bibtitle{Beyond NED: Fast and Effective Search Space Reduction for Complex Question Answering over Knowledge Bases.}
In \bibconf{WSDM 2022}.



\bibitem{Christmann-Explaignn:SIGIR2023}
\bibauthors{Philipp Christmann, Rishiraj Saha Roy, Gerhard Weikum.}
\bibtitle{Explainable Conversational Question Answering over Heterogeneous Sources via Iterative Graph Neural Networks.}
In \bibconf{SIGIR 2023}.

\bibitem{Christmann-CompMix:WWW2024}
\bibauthors{Philipp Christmann, Rishiraj Saha Roy, Gerhard Weikum.}
\bibtitle{CompMix: A Benchmark for Heterogeneous Question Answering.}
In \bibconf{WWW 2024}.

\bibitem{Dejean:arxiv2024}
\bibauthors{Herve Dejean, Stephane Clinchant, Thibault Formal.}
\bibtitle{A Thorough Comparison of Cross-Encoders and LLMs for Reranking SPLADE.}
In \bibconf{arXiv 2024}.

\bibitem{Dhingra-time-aware-LLM:TACL2022}
\bibauthors{Bhuwan Dhingra, Jeremy R Cole, Julian Martin Eisenschlos, Daniel Gillick, Jacob Eisenstein, and William W Cohen.}
\bibtitle{Time-Aware Language Models as Temporal Knowledge Bases.}
In \bibconf{TACL 2022}.


\bibitem{Dong-Trust:PVLDB2015}
\bibauthors{Xin Luna Dong, Evgeniy Gabrilovich, Kevin Murphy, Van Dang, Wilko Horn, Camillo Lugaresi, Shaohua Sun, Wei Zhang.}
\bibtitle{Knowledge-Based Trust: Estimating the Trustworthiness of Web Sources.}
In \bibconf{PVLDB 2015}.

\bibitem{DBLP:journals/arXiv/abs-2312-10997}
\bibauthors{Yunfan Gao, Yun Xiong, Xinyu Gao, Kangxiang Jia, Jinliu Pan, Yuxi Bi, Yi Dai, Jiawei Sun, Qianyu Guo, Meng Wang, Haofen Wang.}
\bibtitle{Retrieval-Augmented Generation for Large Language Models: A Survey.}
In \bibconf{arXiv 2023}.

\bibitem{Gao-citations:emnlp2023}
\bibauthors{Tianyu Gao, Howard Yen, Jiatong Yu, Danqi Chen.}
\bibtitle{Enabling Large Language Models to Generate Text with Citations.}
In \bibconf{EMNLP 2023}.


\bibitem{Guu-REALM:ICML2020}
\bibauthors{Kelvin Guu, Kenton Lee, Zora Tung, Panupong Pasupat, Ming-Wei Chang.}
\bibtitle{Retrieval Augmented Language Model Pre-Training.}
In \bibconf{ICML 2020}.


\bibitem{DBLP:conf/eacl/IzacardG21}
\bibauthors{Gautier Izacard, Edouard Grave.}
\bibtitle{Leveraging Passage Retrieval with Generative Models for Open Domain Question Answering.}
In \bibconf{EACL 2021}.

\bibitem{Jia-TimeQuestions}
\bibauthors{Zhen Jia, Soumajit Pramanik, Rishiraj Saha Roy, and Gerhard Weikum.}
\bibtitle{Complex Temporal Question Answering on Knowledge Graphs.}
In \bibconf{CIKM 2021}.

\bibitem{Jia-FAITH:WWW2024}
\bibauthors{Zhen Jia, Philipp Christmann, Gerhard Weikum.}
\bibtitle{Faithful Temporal Question Answering over Heterogeneous Sources.}
In \bibconf{WWW 2024}.

\bibitem{DBLP:journals/arXiv/abs-2305-06984}
\bibauthors{Ehsan Kamalloo, Nouha Dziri, Charles L. A. Clarke, Davood Rafiei.}
\bibtitle{Evaluating Open-Domain Question Answering in the Era of Large Language Models.}
In \bibconf{arXiv 2023}.

\bibitem{Kandpal:ICML2023}
\bibauthors{Nikhil Kandpal, Haikang Deng, Adam Roberts, Eric Wallace, Colin Raffel.}
\bibtitle{Large Language Models Struggle to Learn Long-Tail Knowledge.}
In \bibconf{ICML 2023}.



\bibitem{DBLP:conf/emnlp/KarpukhinOMLWEC20}
\bibauthors{Vladimir Karpukhin, Barlas Oguz, Sewon Min, Patrick S. H. Lewis, Ledell Wu, Sergey Edunov, Danqi Chen, Wen-tau Yih.}
\bibtitle{Dense Passage Retrieval for Open-Domain Question Answering.}
In \bibconf{EMNLP 2020}.

\bibitem{Lee-MATTER:ACL2024}
\bibauthors{Dongkyu Lee, Chandana Satya Prakash, Jack FitzGerald, Jens Lehmann.}
\bibtitle{MATTER: Memory-Augmented Transformer Using Heterogeneous Knowledge Sources.}
In \bibconf{ACL 2024}.

\bibitem{DBLP:conf/acl/LewisLGGMLSZ20}
\bibauthors{Mike Lewis, Yinhan Liu, Naman Goyal, Marjan Ghazvininejad, Abdelrahman Mohamed, Omer Levy, Veselin Stoyanov, Luke Zettlemoyer.}
\bibtitle{BART: Denoising Sequence-to-Sequence Pre-training for Natural Language Generation, Translation, and Comprehension.}
In \bibconf{ACL 2020}.

\bibitem{DBLP:conf/nips/LewisPPPKGKLYR020}
\bibauthors{Patrick S. H. Lewis, Ethan Perez, Aleksandra Piktus, Fabio Petroni, Vladimir Karpukhin, Naman Goyal, Heinrich Küttler, Mike Lewis, Wen-tau Yih, Tim Rocktäschel, Sebastian Riedel, Douwe Kiela.}
\bibtitle{Retrieval-Augmented Generation for Knowledge-Intensive NLP Tasks.}
In \bibconf{NeurIPS 2020}.

\bibitem{Lin:MC2021}
\bibauthors{Jimmy Lin, Rodrigo Frassetto Nogueira, Andrew Yates.}
\bibtitle{Pretrained Transformers for Text Ranking: BERT and Beyond.}
In \bibconf{Morgan \& Claypool Publishers 2021}.

\bibitem{Liu-SUQL:NAACL2024}
\bibauthors{Shicheng Liu, Jialiang Xu, Wesley Tjangnaka, Sina J. Semnani, Chen Jie Yu, Monica Lam.}
\bibtitle{SUQL: Conversational Search over Structured and Unstructured Data with Large Language Models.}
In \bibconf{NAACL-HLT 2024}.



\bibitem{Mavi:FnT2024}
\bibauthors{Vaibhav Mavi, Anubhav Jangra, Adam Jatowt.}
\bibtitle{Multi-hop Question Answering.}
In \bibconf{Foundations and Trends in Information Retrieval 2024}.

\bibitem{Minaee-LLM-survey}
\bibauthors{Shervin Minaee, Tomas Mikolov, Narjes Nikzad, Meysam Chenaghlu, Richard}
\bibtitle{Socher, Xavier Amatriain, and Jianfeng Gao.}
Large Language Models: A Survey.
In \bibconf{arXiv 2024}.


\bibitem{Nogueira:arxiv2019}
\bibauthors{Rodrigo Frassetto Nogueira, Kyunghyun Cho.}
\bibtitle{Passage Re-ranking with BERT.}
In \bibconf{arXiv 2019}.

\bibitem{Oguz-UniK-QA:NAACL2022}
\bibauthors{Barlas Oguz, Xilun Chen, Vladimir Karpukhin, Stan Peshterliev, Dmytro Okhonko, Michael Sejr Schlichtkrull, Sonal Gupta, Yashar Mehdad, Scott Yih.}
\bibtitle{UniK-QA: Unified Representations of Structured and Unstructured Knowledge for Open-Domain Question Answering.}
In \bibconf{NAACL-HLT 2022}.

\bibitem{RogersGA:CS2023}
\bibauthors{Anna Rogers, Matt Gardner, Isabelle Augenstein.}
\bibtitle{QA Dataset Explosion: A Taxonomy of NLP Resources for Question Answering and Reading Comprehension.}
In \bibconf{ACM Computing Surveys 2023}.

\bibitem{RoyAnand:MC2021}
\bibauthors{Rishiraj Saha Roy, Avishek Anand.}
\bibtitle{Question Answering for the Curated Web: Tasks and Methods in QA over Knowledge Bases and Text Collections.}
In \bibconf{Synthesis Lectures on Information Concepts, Retrieval, and Services, Morgan \& Claypool Publishers 2021}.

\bibitem{Pramanik-Uniqorn:JWS2024}
\bibauthors{Soumajit Pramanik, Jesujoba Alabi, Rishiraj Saha Roy, Gerhard Weikum.}
\bibtitle{UNIQORN: Unified Question Answering over RDF Knowledge Graphs and Natural Language Text.}
In \bibconf{Journal of Web Semantics 2024}.

\bibitem{Sun-PullNet:EMNLP2019}
\bibauthors{Haitian Sun, Tania Bedrax-Weiss, William W. Cohen.}
\bibtitle{PullNet: Open Domain Question Answering with Iterative Retrieval on Knowledge Bases and Text.}
In \bibconf{EMNLP/IJCNLP 2019}.



\bibitem{Sun:NAACL2024}
\bibauthors{Kai Sun, Yifan Ethan Xu, Hanwen Zha, Yue Liu, Xin Luna Dong.}
\bibtitle{Head-to-Tail: How Knowledgeable are Large Language Models (LLMs)? A.K.A. Will LLMs Replace Knowledge Graphs?}
In \bibconf{NAACL-HLT 2024}.


\bibitem{Touvron-LLaMA}
\bibauthors{Hugo Touvron, Thibaut Lavril, Gautier Izacard, Xavier Martinet, Marie-Anne Lachaux, Timothée Lacroix, Baptiste Rozière, Naman Goyal, Eric Hambro, Faisal Azhar, Aurelien Rodriguez, Armand Joulin, Edouard Grave, Guillaume Lample.}
\bibtitle{Llama: Open and efficient foundation language models.}
In \bibconf{arXiv 2023}.

\bibitem{Wu-STARK:arxiv2024}
\bibauthors{Shirley Wu, Shiyu Zhao, Michihiro Yasunaga, Kexin Huang, Kaidi Cao, Qian Huang, Vassilis N. Ioannidis, Karthik Subbian, James Zou, Jure Leskovec.}
\bibtitle{STaRK: Benchmarking LLM Retrieval on Textual and Relational Knowledge Bases.}
In \bibconf{arXiv 2024}.

\bibitem{Wu:IEEE2021}
\bibauthors{Zonghan Wu, Shirui Pan, Fengwen Chen, Guodong Long, Chengqi Zhang, Philip S. Yu.}
\bibtitle{A Comprehensive Survey on Graph Neural Networks.}
In \bibconf{IEEE Transactions on Neural Networks and Learning Systems 2021}.

\bibitem{Yang-CRAG}
\bibauthors{Xiao Yang, Kai Sun, Hao Xin, Yushi Sun, Nikita Bhalla, Xiangsen Chen, Sajal Choudhary, Rongze D. Gui, Ziran W. Jiang, Ziyu Jiang, Lingkun Kong, Brian Moran, Jiaqi Wang, Yifan Ethan Xu, An Yan, Chenyu Yang, Eting Yuan, Hanwen Zha, Nan Tang, Lei Chen, Nicolas Scheffer, Yue Liu, Nirav Shah, Rakesh Wanga, Anuj Kumar, Wen-tau Yih, Xin Luna Dong.}
\bibtitle{CRAG -- Comprehensive RAG Benchmark.}
In \bibconf{arXiv 2024}.

\bibitem{Yasunaga:NAACL2021}
\bibauthors{Michihiro Yasunaga, Hongyu Ren, Antoine Bosselut, Percy Liang, Jure Leskovec.}
\bibtitle{QA-GNN: Reasoning with Language Models and Knowledge Graphs for Question Answering.}
In \bibconf{NAACL-HLT 2021}.


\bibitem{Zhang-Spaghetti:ACL2024}
\bibauthors{Heidi C. Zhang, Sina J. Semnani, Farhad Ghassemi, Jialiang Xu, Shicheng Liu, Monica S. Lam.}
\bibtitle{SPAGHETTI: Open-Domain Question Answering from Heterogeneous Data Sources with Retrieval and Semantic Parsing.}
In \bibconf{ACL 2024}.

\bibitem{Zhang:NAACL2024}
\bibauthors{Jiahao Zhang, Haiyang Zhang, Dongmei Zhang, Yong Liu, Shen Huang.}
\bibtitle{End-to-End Beam Retrieval for Multi-Hop Question Answering.}
In \bibconf{NAACL-HLT 2024}.


\bibitem{Zhao-LLMsurvey}
\bibauthors{Wayne Xin Zhao, Kun Zhou, Junyi Li, Tianyi Tang, Xiaolei Wang, Yupeng Hou, Yingqian Min, Beichen Zhang, Junjie Zhang, Zican Dong, Yifan Du, Chen Yang, Yushuo Chen, Zhipeng Chen, Jinhao Jiang, Ruiyang Ren, Yifan Li, Xinyu Tang, Zikang Liu, Peiyu Liu, Jian-Yun Nie, Ji-Rong Wen.}
\bibtitle{A Survey of Large Language Models.}
In \bibconf{arXiv 2023}.

\bibitem{Zhao:arxiv2024}
\bibauthors{Penghao Zhao, Hailin Zhang, Qinhan Yu, Zhengren Wang, Yunteng Geng, Fangcheng Fu, Ling Yang, Wentao Zhang, Bin Cui.}
\bibtitle{Retrieval-Augmented Generation for AI-Generated Content: A Survey.}
In \bibconf{arXiv 2024}.



\bibitem{Zhu-ODQA-survey:arxiv2021}
\bibauthors{Fengbin Zhu, Wenqiang Lei, Chao Wang, Jianming Zheng, Soujanya Poria, Tat-Seng Chua.}
\bibtitle{Retrieving and Reading: A Comprehensive Survey on Open-domain Question Answering.}
In \bibconf{arXiv 2021}.





\end{thebibliography}
